\documentclass[11pt,a4paper]{article}
\usepackage[hyperref]{acl2020}
\usepackage{times}
\usepackage{latexsym}

\usepackage{graphicx}
\usepackage{colortbl}
\usepackage{color}
\usepackage{float}
\usepackage{amsmath}
\usepackage[mathscr]{eucal}
\usepackage{paralist}

\aclfinalcopy
\usepackage{microtype}

\title{Code-switching patterns can be an effective route\\ to improve performance of downstream NLP applications:\\ A case study of humour, sarcasm and hate speech detection}

\author{Srijan Bansal$^1$, Vishal Garimella$^2$, Ayush Suhane$^3$, Jasabanta Patro$^4$, Animesh Mukherjee$^5$\\
Indian Institute of Technology, Kharagpur, West Bengal, India - 721302\\
\{$^1$srijanbansal97, $^2$vishal\_g,
$^3$ayushsuhane99,
$^4$jasabantapatro,\}@iitkgp.ac.in,\\ $^5$animeshm@cse.iitkgp.ac.in \\
}

\date{}

\begin{document}
\maketitle
\begin{abstract}
In this paper we demonstrate how code-switching patterns can be utilised to improve various downstream NLP applications. In particular, we encode different switching features to improve humour, sarcasm and hate speech detection tasks. We believe that this simple linguistic observation can also be potentially helpful in improving other similar NLP applications. 
\end{abstract}

\section{Introduction}

Code-mixing/switching in social media has become commonplace. Over the past few years, the NLP research community has in fact started to vigorously investigate various properties of such code-switched posts to build downstream applications. The author in \cite{Hidayat2012ANAO} demonstrated that inter-sentential switching is preferred more than intra-sentential switching by Facebook users. Further while 45\% of the switching was done for real lexical needs, 40\% was for discussing a particular topic and 5\% for content classification. In another study~\cite{dey2014} interviewed Hindi-English bilingual students and reported that 67\% of the words were in Hindi and 33\% in English. Recently, many down stream applications have been designed for code-mixed text. \cite{han-etal-2012-automatically} attempted to construct a normalisation dictionary offline using the distributional similarity of tokens plus their string edit distance.  \cite{VyasPOS} developed a POS tagging framework for Hindi-English data.  

More nuanced applications like humour detection~\cite{Humour}, sarcasm detection~\cite{Sarcasm} and hate speech detection~\cite{Hate} have been targeted for code-switched data in the last two to three years.

\subsection{Motivation}
The primary motivation for the current work is derived from~\cite{Vizcano2011AssociationHI} where the author notes -- ``The switch itself may be the object of humour''. In fact,~\cite{siegel_1995} has studied humour in the Fijian language and notes that when trying to be comical, or convey humour, speakers switch from Fijian to Hindi. Therefore, humour here is produced by the \textit{change of code} rather than by the \textit{referential meaning or content of the message}. The paper also talks about similar phenomena observed in Spanish-English cases.

In a study of English-Hindi code-switching and swearing patterns on social networks~\cite{7945452}, the authors show that when people code-switch, there is a strong preference for swearing in the dominant language. These studies together lead us to hypothesize that the patterns of switching might be useful in building various NLP applications.

 \subsection{The present work}
 To corroborate our hypothesis, in this paper, we consider three downstream applications -- (i) humour detection~\cite{Humour}, (ii) sarcasm detection~\cite{Sarcasm} and (iii) hate speech detection~\cite{Hate} for Hindi-English code-switched data. We first provide empirical evidence that the switching patterns between native (Hindi) and foreign (English) words distinguish the two classes of the post, i.e., humour vs non-humour or sarcastic vs non-sarcastic or hateful vs non-hateful. We then featurise these patterns and pump them in the state-of-the-art classification models to show the benefits. We obtain a macro-F1 improvement of 2.62\%, 1.85\% and 3.36\% over the baselines on the tasks of humour detection, sarcasm detection and hate speech detection respectively. As a next step, we introduce a modern deep neural model (HAN - Hierarchical Attention Network \cite{HAN}) to improve the performance of the models further. Finally, we concatenate the switching features in the last hidden layer of the HAN and pass it to the softmax layer for classification. This final architecture allows us to obtain a macro-F1 improvement of 4.9\%, 4.7\% and 17.7\% over the original baselines on the tasks of humour detection, sarcasm detection and hate speech detection respectively.


\section{Dataset}
\label{sec:length}
We consider three datasets consisting of Hindi (hi) - English (en) code-mixed tweets scraped from Twitter for our experiments - Humour, Sarcasm and Hate. We discuss the details of each of these datasets below.

\begin{table}[H]
\centering
\resizebox{\columnwidth}{!}{%
\begin{tabular}{|cccccc|}
\hline
& \textbf{+} & \textbf{-} & \textbf{Tweets} & \textbf{Tokens} & \textbf{Switching*}   \\ 
\hline
\textbf{Humour} & 1755 & 1698 & 3453 & 9851 & 2.20  \\
\hline
\textbf{Sarcasm} & 504 & 4746 & 5250 & 14930 & 2.13\\
\hline
\textbf{Hate}  & 1661 & 2914 & 4575 & 10453 & 4.34\\ 
\hline
\end{tabular}}
\caption{\label{dataset}Dataset description (* denotes average/tweet).}
\end{table}

\noindent\textbf{Humour}: Humour dataset was released by \cite{Humour} and has Hindi-English code-mixed tweets from domains like `sports', `politics', `entertainment' etc. The dataset has uniform distribution of tweets in each category to yield better supervised classification results (see Table~\ref{dataset}) as described by \cite{Du_Balance}. \if{0}A representative example of humorous tweet is \textcolor{red}{\textit{auto}\_en} \textcolor{blue}{\textit{wala}\_hi}
\textcolor{red}{\textit{traffic}\_en}
\textcolor{blue}{\textit{ko}\_hi}
\textcolor{blue}{\textit{itni}\_hi}
\textcolor{blue}{\textit{gaali}\_hi}
\textcolor{blue}{\textit{de}\_hi}
\textcolor{blue}{\textit{raha}\_hi}
\textcolor{blue}{\textit{hai}\_hi}
\textcolor{blue}{\textit{jaise}\_hi}
\textcolor{blue}{\textit{aaj}\_hi}
\textcolor{blue}{\textit{hi}\_hi}
\textcolor{red}{\textit{europe}\_en}
\textcolor{blue}{\textit{se}\_hi}
\textcolor{blue}{\textit{aaya}\_hi}
\textcolor{blue}{\textit{ho}\_hi}\footnote{Gloss: The auto driver is cursing the traffic in a way as if he has come down from Europe just today.}.\fi Here the positive class refers to humorous tweets while the negative class corresponds to non-humorous tweet. Some representative examples from the data showing the point of switch corresponding to the start and the end of the humour component.
\begin{compactitem}
	\item women can crib on things like $\underline{humour_{start}}$ bhaiyya ye shakkar bahot zyada meethi hai $\underline{humour_{end}}$, koi aur quality dikhao\footnote{Gloss: women can crib on things like \textit{brother the sugar is a little more sweet, show a different quality}.}
	\item shashi kapoor trending on mothersday how apt, $\underline{humour_{start}}$ mere paas ma hai $\underline{humour_{end}}$\footnote{Gloss: shashi kapoor trending on mothersday how apt, \textit{I have my mother with me}.}
	\item political journey of kejriwal, from $\underline{humour_{start}}$ mujhe chahiye swaraj $\underline{humour_{end}}$ to $\underline{humour_{start}}$ mujhe chahiye laluraj $\underline{humour_{end}}$\footnote{Gloss: political journey of kejriwal, from \textit{I want swaraj} to \textit{I want laluraj}.}
\end{compactitem}

\noindent\textbf{Sarcasm}: Sarcasm dataset released by~\cite{Sarcasm} contains tweets that have hashtags \#sarcasm and \#irony. Authors used other keywords such as `bollywood', `cricket' and `politics' to collect sarcastic tweets from these domains. In this case, the dataset is heavily unbalanced (see Table~\ref{dataset}). \if{0}A representative example of sarcastic tweet is \textcolor{blue}{\textit{matti}\_hi}
\textcolor{blue}{\textit{say}\_hi}
\textcolor{blue}{\textit{bana}\_hi}
\textcolor{blue}{\textit{hua}\_hi}
\textcolor{blue}{\textit{insan}\_hi}
\textcolor{blue}{\textit{aj}\_hi}
\textcolor{red}{\textit{dust}\_en} \textcolor{red}{\textit{allergy}\_en}
\textcolor{blue}{\textit{ka}\_hi}
\textcolor{blue}{\textit{shikar}\_hi}
\textcolor{blue}{\textit{ha}\_hi} \textcolor{black}{\textit{:P}\_rest}\footnote{Gloss: Clay made man is a victim of dust allergy today}.\fi Here the positive class refers to sarcastic tweets and the negative class means non-sarcastic tweets. Some representative examples from our data showing the point where the sarcasm starts and ends.

\begin{compactitem}
	\item said aib filthy pandit ji, $\underline{sarcasm_{start}}$ aap jo bol rahe ho woh kya shuddh sanskrit hai $\underline{sarcasm_{end}}$? irony shameonyou\footnote{Gloss: said aib filthy pandit ji, \textit{whatever you are telling is it pure sanskrit}? irony shameonyou.} 
	\item irony bappi lahiri sings $\underline{sarcasm_{start}}$ sona nahi chandi nahi yaar toh mila arre pyaar kar le $\underline{sarcasm_{end}}$\footnote{irony bappi lahiri sings \textit{Gloss: doesn't matter you do not get gold or silver, you have got a friend to love}.}
\end{compactitem}

\noindent\textbf{Hate speech}:~\cite{Hate} created the corpus using the tweets posted online in the last five years which have a good propensity to contain hate speech (see Table~\ref{dataset}). Authors mined tweets by selecting certain hashtags and keywords from `politics', `public protests', `riots' etc. The positive class refers to a hateful tweets while the negative class means non-hateful tweets\footnote{The dataset released by this paper only had the hate/non-hate tags for each tweet. However, the language tag for each word required for our experiments was not available. Two of the authors independently language tagged the data and obtained an agreement of 98.1\%. While language tagging, we noted that the dataset is a mixed bag including hate speech, offensive and abusive tweets which have already been shown to be different in earlier works \cite{waseem-etal-2017-understanding}. However, this was the only Hindi-English code-mixed hate speech dataset available.}.An example of 
hate tweet showing the point of switch corresponding to the start and the end of the hate component.
\begin{compactitem}
	\item I hate my university, $\underline{hate_{start}}$ koi us jagah ko aag laga dey $\underline{hate_{end}}$ \footnote{Gloss: I hate my university. \textit{Someone burn that place}.}. 
\end{compactitem}

\if{0}{\color{red}
Our datasets also indicate towards a correlation between code-switching patterns and outcomes like Sarcasm & Humour.Some examples are listed below. [BH: begin humour, EH: end humour, BS: begin sarcasm, ES: end sarcasm]}\fi
\section{Switching features}
In this section, we outline the key contribution of this work. In particular, we identify how patterns of switching correlate with the tweet text being humorous, sarcastic or hateful. We outline a synopsis of our investigation below. 

\subsection{Switching and NLP tasks} 
In this section, we identify how switching behavior is related to the three NLP tasks at our hand.  Let $\mathcal{Q}$ be the property that a sentence has en words which are surrounded by hi words, that is there exists an English word in a Hindi context. For instance, the tweet \textcolor{blue}{\textit{koi}\_hi}
\textcolor{blue}{\textit{to}\_hi}
\textcolor{red}{\textit{pray}\_en}
\textcolor{blue}{\textit{karo}\_hi}
\textcolor{blue}{\textit{mere}\_hi}
\textcolor{blue}{\textit{liye}\_hi}
\textcolor{blue}{\textit{bhi}\_hi} satisfies the property $\mathcal{Q}$. However, \textcolor{blue}{\textit{bumrah}\_hi} \textcolor{blue}{\textit{dono}\_hi} \textcolor{blue}{\textit{wicketo}\_hi} \textcolor{blue}{\textit{ke}\_hi} \textcolor{blue}{\textit{beech}\_hi} \textcolor{blue}{\textit{gumrah}\_hi} \textcolor{blue}{\textit{ho}\_hi} \textcolor{blue}{\textit{gaya}\_hi} does not satisfy $\mathcal{Q}$.

We performed a statistical analysis to determine the correlation between the switching patterns and a classification task at hand (represented by $\mathcal{T}$). Let us denote the probability that a tweet belongs to a positive class for a task $\mathcal{T}$ given that it satisfies property $\mathcal{Q}$ by $p(\mathcal{T}|\mathcal{Q})$. Similarly, let $p(\mathcal{T}|\sim\mathcal{Q})$ be the probability that the tweet belongs to the positive class for task $\mathcal{T}$ and does not satisfy the property $\mathcal{Q}$.

Further let $avg(\mathcal{S}|\mathcal{T})$ be the average switching in positive samples for the task $\mathcal{T}$ and $avg(\mathcal{S}|\sim\mathcal{T})$ denote the average switching in negative samples for the task $\mathcal{T}$.

\begin{table}[H]
    \centering
    \scriptsize
    \resizebox{\columnwidth}{!}{%
    \begin{tabular}{| c | c | c | c |}
         \hline
         &$\mathcal{T}$ : Humour& $\mathcal{T}$ : Sarcasm&$\mathcal{T}$ : Hate\\
         \hline
         $p(\mathcal{T}|\mathcal{Q})$ & 0.56 & 0.28 & 0.36\\
         $p(\mathcal{T}|\sim\mathcal{Q})$ & 0.50 & 0.42 & 0.41\\
         $avg(\mathcal{S}|\mathcal{T})$ & 7.84 & 0.60 & 1.49\\
         $avg(\mathcal{S}|\sim\mathcal{T})$ & 6.50 & 0.89 & 1.54\\
         \hline
    \end{tabular}}
    \caption{Correlation of switching with different classification tasks.}
    \label{table:humourswitch}
\end{table}

The main observations from this analysis for the three tasks -- humour, sarcasm and hate are noted in Table~\ref{table:humourswitch}. For the humour task, $p(humour|\mathcal{Q})$ dominates over $p(humour|\sim\mathcal{Q})$. Further the average number of switching for the positive samples in the humour task is larger than the average number of switching for the negative samples. Finally, we observe a positive Pearson's correlation coefficient of 0.04 between a text being humorous and the text having the property $\mathcal{Q}$. This together indicates that the switching behavior has a positive connection with a tweet being humorous.

On the other hand $p(sarcasm|\sim\mathcal{Q})$ as well as $p(hate|\sim\mathcal{Q})$ respectively dominate over $p(sarcasm|\mathcal{Q})$ and $p(hate|\mathcal{Q})$. Moreover the average number of switching for the negative samples for both these tasks is larger than the average number of switching for the positive samples. The Pearson's correlation between a text being sarcastic (hateful) and the text having the property $\mathcal{Q}$ is negative: -0.17 (-0.04). This shows there is an overall negative connection between the switching behavior and sarcasm/hate speech detection tasks. While we have tested on one language pair (Hindi-English), our hypothesis is generic and has been already noted by linguists earlier~\cite{Vizcano2011AssociationHI}. 

\if{0}As a proof of concept we list some examples in other language-pair tweets below. [BH: begin humour, EH: end humour, BS: begin sarcasm, ES: end sarcasm]
\begin{enumerate}
	\item Don’t forget [BH]la toalla cuando go to le playa[EH] (Spanish) (‘Don’t forget the towel when you go to the beach’)
	\item A dominican goes to buy a soda for 75 cents, he puts in 65 cents & the machine reads [BS]dime[ES] he gets closer & whispers [BS]queiro pepsi[ES] (Spanish)
	\item Cinema [BH]vachinate  telidu  aapude[EH]  success meet  aa (Telugu)
	\item Votes kosam quality leni liquor bottles supply chesina politicians [BS]ee roju quality gurinchi matladuthunnaru[ES] (Telugu)
\end{enumerate}
\fi
\subsection{Construction of the feature vector} 
Motivated by the observations in the previous section we construct a vector \textbf{hi\_en}$[i]$ that denotes the number of Hindi (hi) words before the $i^\textrm{th}$ English (en) word and a vector \textbf{en\_hi}$[i]$ that denotes the number of English (en) words before the $i^\textrm{th}$ Hindi (hi) word. This can also be interpreted as the run-lengths of the Hindi and the English words in the code-mixed tweets. 
Based on these vectors we define nine different features that capture the switching patterns in the code-mixed tweets\footnote{We tried with different other variants but empirically observe that these nine features already subsumes all the necessary distinguishing qualities.}. 

\begin{table}[]
    \centering
    \scriptsize
    \resizebox{\columnwidth}{!}{%
    \begin{tabular}{cl}
         \hline
         Feature name & Description\\
         \hline
         en\_hi\_switches& The number of en to hi switches in a sentence \\
         hi\_en\_switches& The number of hi to en switches in a sentence \\
         V& The total number of switches in a sentence  \\
         fraction\_en & Fraction of English words in a sentence \\
         fraction\_hi & Fraction of Hindi words in a sentence \\
         mean  hi\_en & Mean of hi\_en vector \\
         stddev hi\_en  & Standard deviation of hi\_en vector \\ 
         mean   en\_hi & Mean of en\_hi vector \\
         stddev en\_hi & Standard deviation of en\_hi vector \\
         \hline
    \end{tabular}}
    \caption{\label{table:features1}Description of the switching features.}
\end{table}


\noindent\textbf{An example feature vector computation}: Consider the sentence - \textcolor{blue}{\textit{koi}\_hi} \textcolor{blue}{\textit{to}\_hi} \textcolor{red}{\textit{pray}\_en} \textcolor{blue}{\textit{karo}\_hi} \textcolor{blue}{\textit{mere}\_hi} \textcolor{blue}{\textit{liye}\_hi} \textcolor{blue}{\textit{bhi}\_hi}.\\
\textbf{hi\_en}\hspace{44pt}: $[0, 0, 2, 0, 0, 0, 0]$ \\
\textbf{en\_hi}\hspace{41pt} : $[0, 0, 0, 1, 1, 1, 1]$
\\ 
\textbf{Feature vector}: $[1, 1, 2, {\frac{1}{7}}, {\frac{6}{7}}, {\frac{2}{7}}, 0.69, {\frac{4}{7}}, 0.49]$ \\ 

\section{Experiments}
 \subsection{Pre-processing}
 Tweets are tokenized and punctuation marks are removed. All the hashtags, mentions and urls are  stored and converted to string `hashtag', `mention' and `url' to capture the general semantics of the tweet. Camel-case hashtags were segregated and included in the tokenized tweets (see~\cite{NoisyText1},~\cite{NoisyText2}). For example, \textit{\#AadabArzHai} can be decomposed into  three distinct words: \textit{Aadab}, \textit{Arz} and \textit{Hai}. We use the same pre-processing for all the results presented in this paper.

\subsection{Machine learning baselines}
\noindent{\bf Humour baseline}~\cite{Humour}: Uses features such as $n$-grams, bag-of-words, common words and hashtags to train the standard machine learning models such as SVM and Random-Forest. The authors used character $n$-grams, as previous work shows that this feature is very efficient in classifying text because they do not require expensive text pre-processing techniques like tokenization, stemming and stop words removal. They are also language independent and can be used in code-mixed texts. In their paper, the authors report the results for tri-grams.

\noindent{\bf Sarcasm baseline}~\cite{Sarcasm}: This model also uses a combination of word $n$-grams, character $n$-grams, presence or absence of certain emoticons and sarcasm indicative tokens as features. A sarcasm indicative score is computed
and chi-squared feature reduction is used to take the top 500 most relevant words. These were incorporated into features used for classification. Standard off-the-shelf machine learning models like SVM and Random Forest were used. 

\noindent{\bf Hate baseline}~\cite{Hate}: The hate speech detection baseline also consists of similar features such as character $n$-grams, word $n$-grams, negation words \footnote{see Christopher Pott’s sentiment tutorial: \url{http://sentiment.christopherpotts.net/lingstruc.html}} and a lexicon of hate indicative tokens. Chi-squared feature reduction method was used to decrease the dimensionality of the features. Once again SVM and Random Forest based classifiers were used for this task.


\noindent\textbf{Switching features}: We plug in the nine switching features introduced in the previous section to the three baseline models for humour, sarcasm and hate speech detection. 


\subsection{Deep learning architecture}

In order to draw the benefits of the modern deep learning machinery, we build an end-to-end model for the three tasks at hand. We use the Hierarchical Attention Network (HAN) \cite{HAN} which is one of the state-of-the-art models for text and document classification. It can represent sentences in different levels of granularity by stacking recurrent neural networks on character, word and sentence level by attending over the words which are informative.
We use the GRU implementation of HAN to encode the text representation for all the three tasks. 

\noindent\textbf{Handling data imbalance by sub-sampling}: Since the sarcasm dataset is heavily unbalanced we sub-sampled the data to balance the classes. To this purpose, we categorise the negative samples into those that are easy or hard to classify. Hypothesizing that if a model can predict the hard samples reliably it can do the same with the easy samples. We trained a classifier model on the training dataset and obtained the softmax score which represents $p(sarcastic|text)$ for the test samples. Those test samples which have a score less than a very low confidence score (say 0.001) are removed imagining them to be easy samples. The dataset thus got reduced and more balanced. It is important to note that positive samples are never removed. We validated this hypothesis through the test set. Our trained HAN model achieves an accuracy of 94.4\% in classifying the easy (thrown out) samples as non-sarcastic thus justifying the sub-sampling.

\begin{figure} [h!]
    \centering
    \resizebox{\columnwidth}{!}{%
    \includegraphics[height=6.5cm,width=8.5cm]{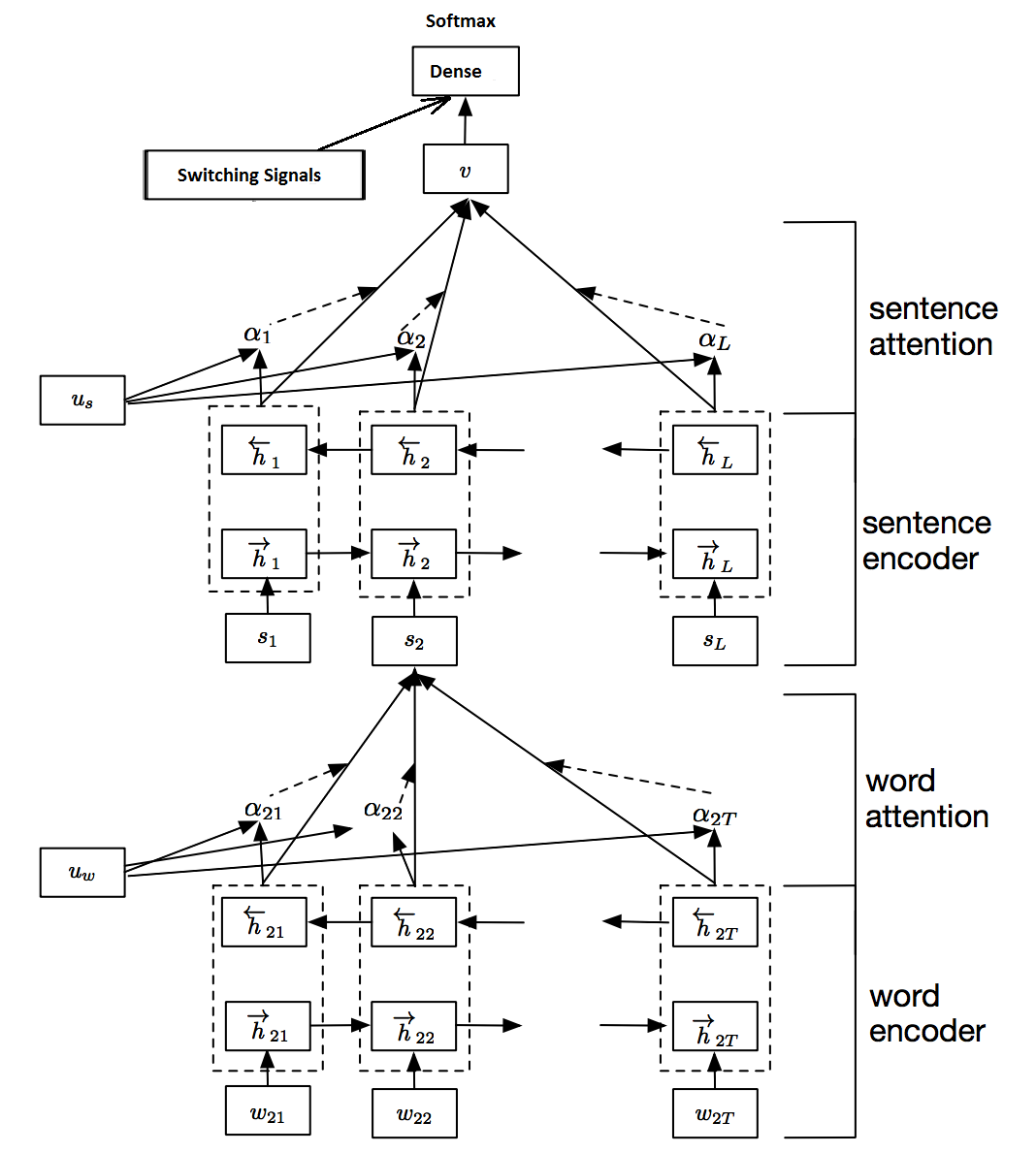}}
    \caption{\label{fig:my_label} The overall HAN architecture along with the switching features in the final layer.}
\end{figure}

\noindent\textbf{Switching features}: We include the switching features to the pre-final fully-connected layer of HAN to observe if this harnesses additional benefits (see Figure~\ref{fig:my_label}). 


\subsection{Experimental Setup}

\noindent\textbf{Train-test split}: For all datasets, we maintain a train-test split of 0.8 - 0.2 and perform 10-fold cross validation.

\noindent\textbf{Parameters of the HAN}: BiLSTMs: no dropout, early stopping patience: 15, optimizer = `adam' (learning rate = 0.001, beta\_1 = 0.9), loss = binary cross entropy, epochs = 200, batch\_size = 32, pre-trained word-embedding size = 50, hidden size: $[20,60]$, dense output size (before concatenation): $[15,30]$.

\noindent\textbf{Pre-trained embeddings}: We obtained pre-trained embeddings by training GloVe from scratch using the large code-mixed dataset (725173 tweets) released by \cite{patro-etal-2017-english} plus all the tweets (13278) in our three datasets.
\section{Results}
We compare the baseline models along with (i) the baseline + switching feature-based models and (ii) the HAN models. We use macro-F1 score for comparison all through. The main results are summarized in Table~\ref{charHAN-table}. The interesting observations that one can make from these results are -- (i) inclusion of the switching features always improves the overall performance of any model (machine learning or deep learning) for all the three tasks, (ii) the deep learning models are always better than the machine learning models. Inclusion of switching features into the machine learning models (indicated as BF in Table~\ref{charHAN-table}) allows us to obtain a macro-F1 improvement of 2.62\%, 1.85\% and 3.36\% over the baselines (indicated as B in Table~\ref{charHAN-table}) on the tasks of humour detection, sarcasm detection and hate speech detection respectively. Inclusion of the switching feature in the HAN model (indicated as HF in Table~\ref{charHAN-table}) allows us to obtain a macro-F1 improvement of 4.9\%, 4.7\% and 17.7\% over the original baselines (indicated as B in Table~\ref{charHAN-table}) on the tasks of humour detection, sarcasm detection and hate speech detection respectively. 

\begin{table}[]
\centering
\scriptsize
\resizebox{\columnwidth}{!}{%
\begin{tabular}{l| c |  c | c }
\hline \textbf{Model}&\textbf{Humour}& \textbf{Sarcasm}& \textbf{Hate} \\ \hline
\cellcolor{red!20}Baseline (B) & 
\cellcolor{red!20}69.34 & 
\cellcolor{red!20}78.4 & 
\cellcolor{red!20}33.60 \\
\cellcolor{red!20}Baseline + Feature (BF) & \cellcolor{red!20}\textbf{71.16} & \cellcolor{red!20}\textbf{79.85} & \cellcolor{red!20}\textbf{34.73}\\
\cellcolor{green!20}HAN (H) & \cellcolor{green!20}72.04 & 
\cellcolor{green!20}81.36 & 
\cellcolor{green!20}38.78 \\
\cellcolor{green!20}HAN + Feature (HF) &  \cellcolor{green!20}\textbf{72.71} & \cellcolor{green!20}\textbf{82.07} & \cellcolor{green!20}\textbf{39.54}\\
\hline
\end{tabular}}
\caption{\label{charHAN-table} Summary of the results from different models in terms of macro-F1 scores. M-W U test shows all improvements of HF over B are significant.}
\end{table}

\noindent\textbf{Success of our model}: Success of our approach is evident from the following examples. For instance, as we had demonstrated earlier, humour is positively correlated with switching, a tweet having a switching pattern like -
\textcolor{blue}{\textit{anurag}\_hi}
\textcolor{blue}{\textit{kashyap}\_hi}
\textcolor{red}{\textit{can}\_en}
\textcolor{red}{\textit{never}\_en}
\textcolor{red}{\textit{join}\_en}
\textcolor{blue}{\textit{aap}\_hi}
\textcolor{red}{\textit{because}\_en}
\textcolor{red}{\textit{ministers}\_en}
\textcolor{red}{\textit{took}\_en}
\textcolor{red}{\textit{oath}\_en},
``\textcolor{blue}{\textit{main}\_hi}
\textcolor{blue}{\textit{kisi}\_hi }
\textcolor{blue}{\textit{anurag}\_hi}
\textcolor{blue}{\textit{aur}\_hi}
\textcolor{blue}{\textit{dwesh}\_hi}
\textcolor{blue}{\textit{ke}\_hi}
\textcolor{blue}{\textit{bina}\_hi}
\textcolor{blue}{\textit{kaam}\_hi}
\textcolor{blue}{\textit{karunga}\_hi}'' which was not detected as humorous by the baseline (B) but was detected so by our models (BF and HF). Note that the author of the above tweet seems to have categorically switched to Hindi to express the humour; such observations have also been made in~\cite{Rudra16} where opinion expression was cited as a reason for switching.

Sarcasm being negatively correlated with switching, a tweet without having switching is more likely to be sarcastic. For instance, the tweet \textcolor{blue}{\textit{naadaan}\_hi} \textcolor{blue}{\textit{baalak}\_hi} \textcolor{blue}{\textit{kalyug}\_hi} \textcolor{blue}{\textit{ka}\_hi} \textcolor{blue}{\textit{vardaan}\_hi} \textcolor{blue}{\textit{hai}\_hi}
\textcolor{blue}{\textit{ye}\_hi}, which bears no switching was labeled non-sarcastic by the baseline. Our models (BF and HF) have rectified it and correctly detected it as sarcastic. 

Similarly, hate being negatively correlated with switching, a tweet with no switching - \textcolor{blue}{\textit{shilpa}\_hi}
\textcolor{blue}{\textit{ji}\_hi}
\textcolor{blue}{\textit{aap}\_hi}
\textcolor{blue}{\textit{ravidubey}\_hi}
\textcolor{blue}{\textit{jaise}\_hi}
\textcolor{blue}{\textit{tuchho}\_hi}
\textcolor{blue}{\textit{ko}\_hi}
\textcolor{blue}{\textit{jawab}\_hi}
\textcolor{blue}{\textit{mat}\_hi}
\textcolor{blue}{\textit{dijiye}\_hi}
\textcolor{blue}{\textit{ye}\_hi}
\textcolor{blue}{\textit{log}\_hi}
\textcolor{blue}{\textit{aap}\_hi}
\textcolor{blue}{\textit{ke}\_hi}
\textcolor{blue}{\textit{sath}\_hi}
\textcolor{blue}{\textit{kabhi}\_hi}
\textcolor{blue}{\textit{nahi}\_hi} was labeled as non-hateful by the baseline, was detected as hateful by our methods (BF and HF).


\section{Conclusion}

In this paper, we identified how switching patterns can be effective in improving three different NLP applications. We present a set of nine features that improve upon the state-of-the-art baselines. In addition, we exploit the modern deep learning machinery to improve the performance further. Finally, this model can be improved further by pumping the switching features in the final layer of the deep network.

In future, we would like to extend this work for other language pairs. For instance, we have seen examples of such switching in English-Spanish\footnote{Don’t forget $\underline{humour_{start}}$ \textit{la toalla cuando go to le playa} $\underline{humour_{end}}$; Gloss: Don’t forget the towel when you go to the beach.} and English-Telugu\footnote{Votes kosam quality leni liquor bottles supply chesina politicians $\underline{sarcasm_{start}}$ \textit{ee roju quality gurinchi matladuthunnaru} $\underline{sarcasm_{end}}$; Gloss:{Politicians who supplied low quality liquor bottles for votes are talking about quality today}.} pairs also. Further we plan to investigate other NLP applications that can benefit from the simple linguistic features introduced here.

\bibliography{acl2020.bib}
\bibliographystyle{acl_natbib.bst}

\end{document}